\documentclass[10pt,twocolumn,letterpaper]{article}

\usepackage{iccv}
\usepackage{times}
\usepackage{epsfig}
\usepackage{graphicx}
\usepackage{amsmath}
\usepackage{amssymb}
\usepackage{multirow}
\usepackage{array}
\usepackage{booktabs}
\usepackage{verbatim}
\usepackage{times}
\usepackage{epsfig}
\usepackage{amsmath}
\usepackage{amssymb}
\usepackage{enumitem}
\usepackage{algorithm}
\usepackage{algorithmic}
\usepackage{bm}
\usepackage{arydshln}
\usepackage{makecell}
\usepackage{setspace}
\usepackage{threeparttable}
\usepackage{subfigure}
\usepackage{threeparttable}

\usepackage[pagebackref=true,breaklinks=true,letterpaper=true,colorlinks,bookmarks=false]{hyperref}

\newcommand{\urllink}{\fontsize{7.5pt}{\baselineskip}\selectfont}

\iccvfinalcopy

\ificcvfinal\fi

\begin{document}

\title{Differentiable Convolution Search for Point Cloud Processing}

\author{
Xing Nie$^{1,2}$,
Yongcheng Liu$^{1}$,
Shaohong Chen$^4$,
Jianlong Chang$^3$, \\
Chunlei Huo$^1$\thanks{Corresponding author.} ,
Gaofeng Meng$^{1,2,5}$,
Qi Tian$^3$,
Weiming Hu$^1$,
Chunhong Pan$^1$,
\\[0.2cm]
\small$ ^1$ National Laboratory of Pattern Recognition, Institute of Automation, Chinese Academy of Sciences. \\
\small$ ^2$ School of Artificial Intelligence, University of Chinese Academy of Sciences.
\small$ ^3$ Huawei Cloud \& AI. \\
\small$ ^4$ Xidian University.
\small$ ^5$ Centre for Artificial Intelligence and Robotics, HK Institute of Science \& Innovation, CAS.\\
\small\texttt{Email:\;niexing2019@ia.ac.cn,\;\{yongcheng.liu,\;clhuo\}@nlpr.ia.ac.cn} \\
}

\maketitle
\ificcvfinal\thispagestyle{empty}\fi

\begin{abstract}
Exploiting convolutional neural networks for point cloud processing is quite challenging, due to the inherent irregular distribution and discrete shape representation of point clouds.
To address these problems, many handcrafted convolution variants have sprung up in recent years.
Though with elaborate design, these variants could be far from optimal in sufficiently capturing diverse shapes formed by discrete points.
In this paper, we propose PointSeaConv, \textit{i.e.}, a novel differential convolution search paradigm on point clouds.
It can work in a purely data-driven manner and thus is capable of auto-creating a group of suitable convolutions for geometric shape modeling.
We also propose a joint optimization framework for simultaneous search of internal convolution and external architecture, and introduce epsilon-greedy algorithm to alleviate the effect of discretization error.
As a result, PointSeaNet, a deep network that is sufficient to capture geometric shapes at both convolution level and architecture level, can be searched out for point cloud processing.
Extensive experiments strongly evidence that our proposed PointSeaNet surpasses current handcrafted deep models on challenging benchmarks across multiple tasks with remarkable margins.
\end{abstract}

\section{Introduction}

Recently, 3D point cloud processing has received great attention, since it plays an important role in the fields of autonomous driving, robotics, geomatics, and so on. Nevertheless, compared with 2D image processing, this task is quite challenging due to the non-grid structure and orderless permutation of point clouds. Furthermore, it is extremely difficult to perform shape analysis for point clouds, as the underlying shape formed by those discrete points is visually elusive to capture.

In order to tackle this task, many geometric descriptors~\cite{tombari2010unique} have been manually designed over the past decades. Though usable in certain scenarios, these descriptors often suffer from unsatisfactory performance and poor task-adaptivity. Recently, convolutional neural networks (CNN) in deep learning technologies has made remarkable achievements in the processing of regular data, \textit{e.g.}, image \cite{krizhevsky2012imagenet, ren2015faster}, video \cite{simonyan2014very}, and speech \cite{oord2016wavenet}. Accordingly,  there has been a growing interest in exploiting the power of CNN for irregular point cloud processing \cite{liu2019relation, qi2017pointnet++, wang2019dynamic}.

\begin{figure}[t]
\centering
\includegraphics[width=0.48\textwidth]{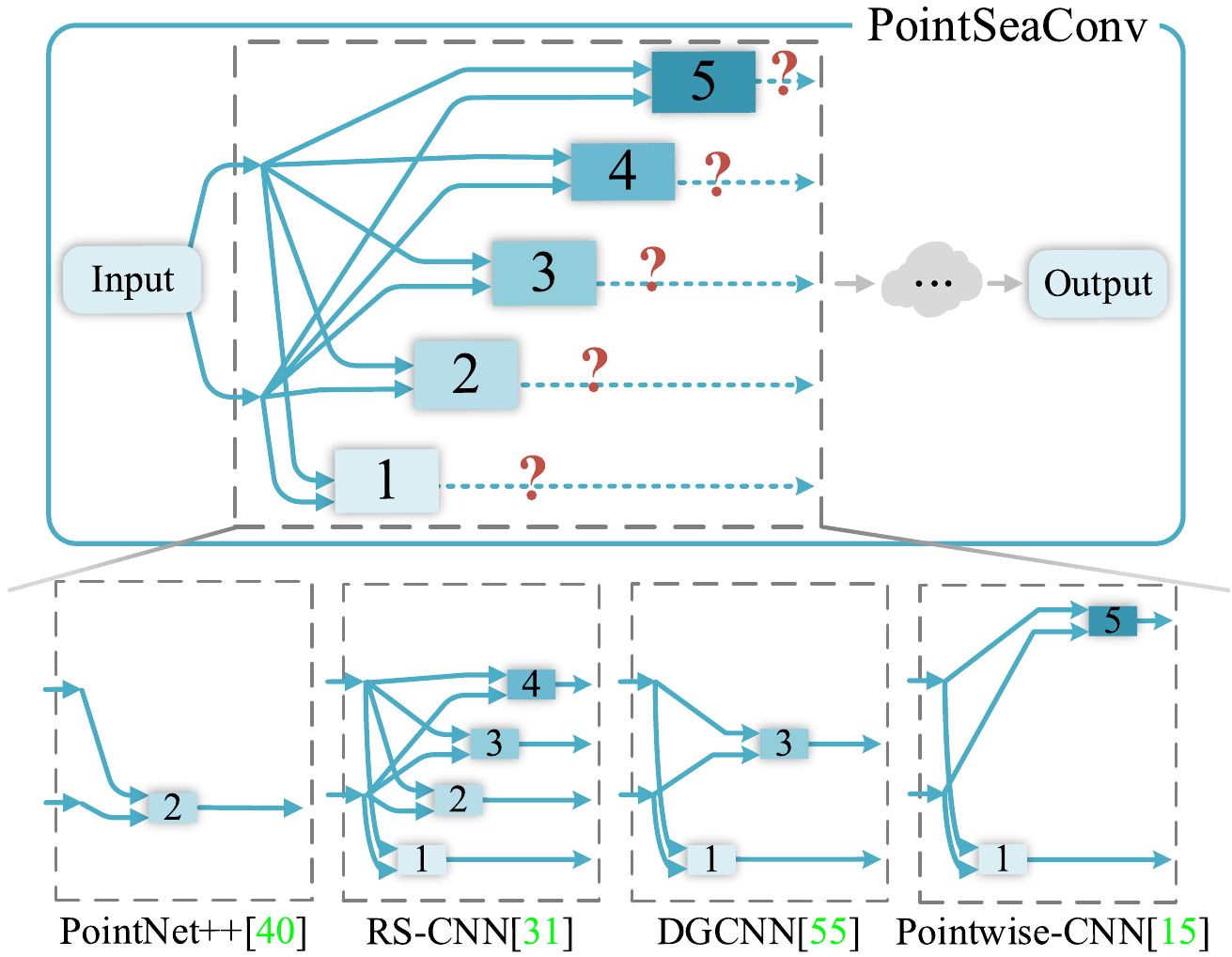}
\vspace{-4mm}
\caption{A sketch of the proposed convolution search paradigm, \textit{i.e.}, PointSeaConv. PointSeaConv is achieved by constructing a dynamic and learnable directed acyclic graph (DAG). Most handcrafted convolutions can be regarded as its special cases.}
\label{Fig1}
\vspace{-0.5cm}
\end{figure}

To facilitate the application of classic CNN, some researchers transform point clouds into regular multi-view images~\cite{guo2020deep, su2015multi} or voxel grids~\cite{qi2019deep, zhou2018voxelnet}. While practicable, these transformations usually lead to the loss of shape information because of self-occlusions or quantization artifacts. As a pioneer, PointNet~\cite{qi2017pointnet} learns directly on point clouds with shared multi-layer perceptrons (MLP) and max-pooling operation. Despite its encouraging performance on shape analysis, PointNet has difficulty in learning fine-grained shape representation due to the lack of local modeling. To overcome this issue, PointNet++~\cite{qi2017pointnet++} builds an explicit local-to-global CNN-like architecture with multiple set abstraction layers. However, it performs convolution by simply applying PointNet on the local regions, which could be powerless in capturing diverse local structures.

Plenty of follow-up research, therefore, is devoted to manually design convolution variants, which is expected to grasp local structures well. The typical methods along this route are EdgeConv~\cite{wang2019dynamic}, PointCNN~\cite{li2018pointcnn}, RS-Conv~\cite{liu2019relation}, KPConv~\cite{thomas2019kpconv}, PointConv~\cite{wu2019pointconv}, and so on. They either construct local graph connections collocated with graph convolution methods, or empirically introduce local geometric statistics~(\textit{e.g.}, density) into convolution operation. Though achieving decent performance, these convolutions greatly depend on heuristic rules and experienced engineering.

In this paper, we argue that the manually-designed convolution could be suboptimal for point cloud processing, especially in the era of data-driven deep learning. The key challenge for convolution learning on point clouds is how to make it being capable to sufficiently capture diverse local structures. This motivates us, accordingly, to construct an auto-created convolution search paradigm, which can be directly driven by irregular structures in point clouds.

To this end, we propose PointSeaConv, \textit{i.e.}, a novel differentiable convolution search paradigm on point clouds. Concretely, we first formulate a general convolution for geometric structure modeling and transform it into a searchable process. This is achieved by constructing a dynamic and learnable directed acyclic graph (DAG). Consequently, the convolution expression can be determined by the DAG while the convolution weight can be learned on the DAG. Moreover, most handcrafted convolutions can be regarded as special cases of our searchable one (Fig.~\ref{Fig1}). We then develop a joint and differentiable optimization framework for optimizing the search of internal convolution and external architecture, simultaneously. Especially, the epsilon-greedy algorithm is introduced into the search process, which greatly alleviates the effect of discretization error. As a result, PointSeaNet, a deep network that captures geometric structures at both convolution level and architecture level, can be searched out for point cloud processing.

The key contributions can be summarized as follows:

\begin{itemize}[leftmargin=2em]
\item We propose a novel differentiable convolution search paradigm, \textit{i.e.}, PointSeaConv. It can work in a purely data-driven manner and thus is capable of creating a group of suitable convolutions for point cloud processing. To our best knowledge, we are the first to conduct fundamental convolution search on point clouds.

\item We propose a joint and differential optimization framework for simultaneous search of internal convolution and external architecture. Under the framework, PointSeaNet, a deep network that sufficiently captures geometric structures of point clouds at both convolution level and architecture level, can be searched out.

\item We innovatively introduce epsilon-greedy algorithm into the search framework. Thanks to the algorithm, the adverse effect of discretization error can be greatly alleviated during the whole search process.

\end{itemize}

\section{Related Work}

\subsection{Point Cloud Processing}

In this section, we briefly review existing deep learning methods for for point cloud processing. According to the data type of input, these methods can be generally divided into projection-based networks and point-based networks.

Projection-based networks~\cite{su2015multi, tatarchenko2018tangent, yang2019learning} project 3D point clouds into 2D multiple views from various angles. Despite of impressive performance, most of them suffer from information loss due to occluded surfaces and viewpoint selections. Alternatively, volumetric-based networks~\cite{graham20183d, maturana2015voxnet, su2018splatnet} convert point clouds into uniform 3D grids and then apply CNNs on the volumetric grids. The key criticisms of these methods are the heavy computational burden and loss of details. Unlike these methods, our work is able to directly process point clouds without any pre/post-processing step.

Point-based networks directly consume point cloud and become increasingly popular. Inspired by PointNet~\cite{qi2017pointnet}, much research has been devoted to elaborately designing sophisticated networks to learn pointwise local features. These methods can be generally classified as 1) pointwise MLP networks~\cite{achlioptas2018learning, hu2020randla, qi2017pointnet++, sun2019srinet, zhao2019pointweb}, 2) point convolution networks~\cite{atzmon2018point, hermosilla2018monte, liu2019relation, thomas2019kpconv, wu2019pointconv}, 3) data indexing networks~\cite{klokov2017escape, lei2019octree, riegler2017octnet, zeng20183dcontextnet}. However, these methods lack internal mechanisms to generate convolution operators according to local geometric structures. In contrast, our PointSeaNet can automatically search fundamental convolution operations driven by input point clouds.

\subsection{Neural Architecture Search (NAS)}

Neural architecture search~(NAS) methods inherently aim to provide an automatic way of designing architectures to replace the manual ones. Early methods employ reinforcement learning~\cite{zoph2016neural, zoph2018learning} and evolutionary algorithm~\cite{chen2019renas, xie2017genetic} to find the optimal architecture. Further, one shot approaches~\cite{bender2018understanding, brock2017smash, cai2018proxylessnas, liu2018darts} are proposed to reduce the computational costs by training the super-network only once, which is sampled and evaluated subsequently. The pioneering work DARTS~\cite{liu2018darts} introduces a differentiable framework to relax the search space and hence improves the efficiency of search period. Most of them are elaborately designed to tackle various 2D vision problems and have achieved superior performance~\cite{chen2019detnas, peng2019efficient}. Recently, some approaches have focused on neural architecture search for irregular point cloud processing~\cite{li2020sgas, tang2020searching, yu2020c2fnas}. However, these methods heavily rely on fixed convolution operators, such as existing graph convolutions (\textit{e.g.}, EdgeConv~\cite{wang2019dynamic}, GAT~\cite{velivckovic2017graph} and SemiGCN~\cite{kipf2016semi}) and pre-defined convolution kernels on 2D images, which results in incapability to sufficiently capture geometric structures for point clouds. Through our differentiable convolution search paradigm, by comparison, fundamental convolution operators collaboratively working with external architecture can be searched out.

\section{Methodology}
\label{method}

In this section, we first formulate a general convolution (Sec.~\ref{section1}), of which the image convolution can be seen as a special case. We then adapt this general convolution to learn geometric information in point clouds, by transforming it into a convolution search problem (Sec.~\ref{section2}). Finally, we show how the convolution search can be collocated with external architecture search in a joint and differentiable optimization manner (Sec.~\ref{section3}).

\subsection{General Convolution Formulation}
\label{section1}

\noindent \textbf{General convolution.}\,\,~The key properties of convolution are local connectivity and weight sharing (over different local regions)~\cite{lecun1998gradient}. Technically, inside a local region, the convolution can be generally decomposed into two steps: (i) transforming the feature vector of each unit in this local region and (ii) aggregating all the transformed features for summarizing the local information. Formally, given a local region $\{\boldsymbol{p}_{1}, \boldsymbol{p}_{2},\ldots, \boldsymbol{p}_{n}\}$ with $n$ units, in which $\boldsymbol{p}_{j} \in \mathbb{R}^{F\times 1}$ denotes the feature vector of the $j$-th unit, then the general convolution with above two steps can be formulated as
\begin{equation}\label{eq:general_conv}
\boldsymbol{p}= \mathcal{G} \big( \{\psi(\boldsymbol{p}_{j}) \}_{j=1,\ldots,n} \big) \footnote{In this paper, we omit the bias term and activation function for clarity.},
\end{equation}
where the output $\boldsymbol{p} \in \mathbb{R}^{F'\times 1}$ is obtained by a transformation function $\psi(\cdot)$ at step (i) and an aggregation function $\mathcal{G(\cdot)}$ at step (ii). In addition, $\psi(\cdot)$ is usually shared over different local regions to achieve weight sharing property.

\vspace{2pt}
\noindent \textbf{Image convolution.}\,\,~Notably, the image convolution can be seen as a special case of this general convolution. To be specific, the local region in the image is arranged with a regular grid structure and all the units (\textit{i.e.}, pixels) in this region are fixed and ordered. Thus the image convolution can be written as
\begin{equation}\label{eq:image_conv}
\boldsymbol{p}=\sum_{j} \boldsymbol{\mathrm{W}}_j \odot \boldsymbol{p}_{j},
\end{equation}
where $\boldsymbol{\mathrm{W}}_j \in \mathbb{R}^{F^{\prime} \times F}$ denotes the convolutional weight matrix for $\boldsymbol{p}_{j}$ and ``$\odot$'' indicates the matrix multiplication. That is, $\psi(\boldsymbol{p}_{j})$ and $\mathcal{G(\cdot)}$ in Eq.~\eqref{eq:general_conv} are implemented as $\boldsymbol{\mathrm{W}}_j \odot \boldsymbol{p}_{j}$ and summation here, respectively. Moreover, note that the weight $\boldsymbol{\mathrm{W}}$ is learned on pixel values, \textit{i.e.}, $\boldsymbol{p}_j|_{j=1,\ldots,n}$, hence it shows great power to capture semantic patterns reflected by color information.

\begin{table}[t]
\footnotesize
\begin{threeparttable}
\resizebox{0.47\textwidth}{!}{
\begin{tabular}{p{3.7cm}ll}
\toprule
Essential association (EA) &  Advantages \\
\midrule
$e_1\text{:}\, \boldsymbol{n}_{i}$ & global features~\cite{qi2017pointnet} \\
$e_2\text{:}\, \boldsymbol{n}_{j}$ & local features~\cite{atzmon2018point, komarichev2019cnn, li2019deepgcns, liu2019relation, qi2017pointnet++} \\
$e_3\text{:}\, \boldsymbol{n}_{i}-\boldsymbol{n}_{j}$ & geometric relations~\cite{hu2020randla, li2018pointcnn, liu2019relation, simonovsky2017dynamic, xu2018spidercnn} \\
$e_4\text{:}\, ||\boldsymbol{n}_{i}-\boldsymbol{n}_{j}||_2$ & Euclidean distance~\cite{hu2020randla, liu2019relation, pan20183dti, tatarchenko2018tangent, thomas2019kpconv} \\
$e_5\text{:}\, \boldsymbol{n}_{i}-\sum_{\boldsymbol{n}_{k} \in \mathcal{N}\left(\boldsymbol{n}_{i}\right)} \frac{\boldsymbol{n}_{k}}{\left|\mathcal{N}\left(\boldsymbol{n}_{i}\right)\right|} $ & salient information~\cite{hua2018pointwise, tatarchenko2018tangent}   \\
\bottomrule
\end{tabular}}
\vspace{0.1cm}
\end{threeparttable}
\caption{A summary of five essential association (EA) candidates. $\left|\mathcal{N}\left(\boldsymbol{n}_{i}\right)\right|$ indicates the number of all points in $\mathcal{N}\left(\boldsymbol{n}_{i}\right)$.}
\vspace{-0.5cm}
\label{tab: EA}
\end{table}

\subsection{Convolution Search on Point Clouds}
\label{section2}
\noindent \textbf{Existing challenges.}\,\,~Recently, the image convolution in Eq.~\eqref{eq:image_conv} has been adapted by many researchers for transferring its great power in image processing into point cloud processing. However, this is in fact quite challenging. The reasons are twofold: (i) it is very intractable to achieve the permutation invariance to point set, whilst ensuring that the convolution is capable of sufficiently learning local structures; (ii) the weight sharing property of convolution is hard to implement due to the irregular structures (\textit{i.e.}, variable number of points) over different local regions. Although these issues are partly alleviated by recent convolution variants~\cite{li2018pointcnn, liu2019relation, qi2017pointnet++, wang2019dynamic, xu2018spidercnn}, most of them are manually designed. Such handcrafted convolutions not only rely heavily on expert knowledge with long-term design cycle, but also show poor generalization in various scenarios \cite{liu2020closer}.

\begin{figure*}[t]
\centering
\includegraphics[width=1\textwidth]{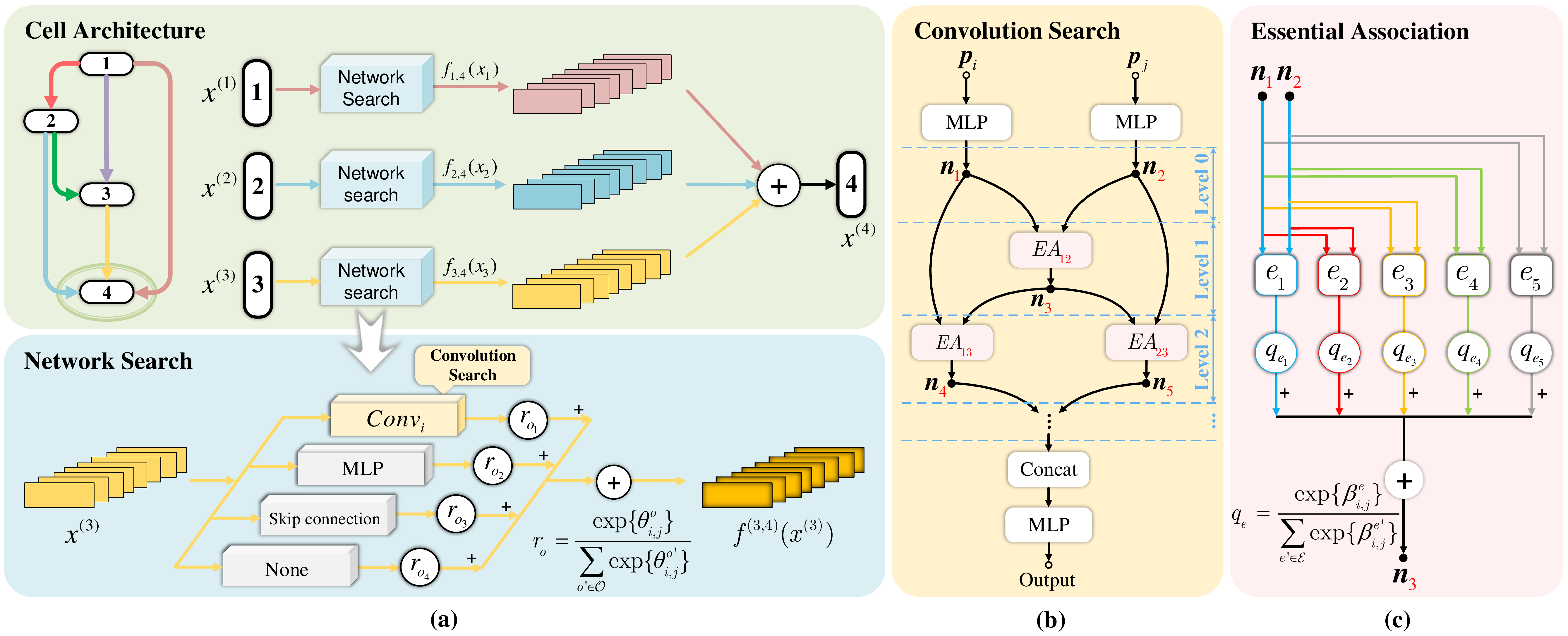}
\vspace{-4mm}
\caption{An overview of our PointSeaNet. For clarity, a cell architecture (the upper part of (a)) formed by four ordered nodes is illustrated. The search process of the whole network can be divided into convolution search and cell architecture search (\textit{i.e.}, network search). The convolution search is achieved under our geometric convolution modeling in Eq.~\eqref{eq:point_conv}, which is transformed from the general convolution in Eq.~\eqref{eq:general_conv}. Technically, it is constructed with a multi-level directed acyclic graph (DAG, Sec.~\ref{DAG}), in which a set of searchable essential associations (EA, Sec.~\ref{EA}) are conducted. The essential association is searched out from five fundamental candidates, \textit{i.e.}, $e_1 \sim e_5$ in Tab.~\ref{tab: EA}. Moreover. the proposed convolution search can be collocated with external network search in a joint and differentiable optimization manner (Sec.~\ref{section3}). As a result, a group of suitable convolutions can be searched out for point cloud processing in a purely data-driven manner. Note that the number of searchable convolutions, \textit{i.e.}, \textit{Conv}$_i$ in network search, is variable. Best viewed in color.}
\vspace{-0.5cm}
\label{Fig3}
\end{figure*}

\vspace{2pt}
\noindent \textbf{Geometric modeling.}\,\,~In this paper, we argue that it could be suboptimal to manually design the convolution for point cloud processing. Hence we propose an entirely different route, \textit{i.e.}, transforming the general convolution in Eq.~\eqref{eq:general_conv} into a convolution search problem on point clouds. Formally, we model the points in a local region as the central point $\boldsymbol{p}_{i}$ and its surrounding neighbors $\boldsymbol{p}_{j} \in \mathcal{N}(\boldsymbol{p}_{i})$. Note that the shape information in point clouds is from relative spatial distribution among points. This is quite different from the 2D image, where meaningful information is from the value of pixels, not grid distribution. Accordingly, we propose to learn the shape information by learning the geometric associations between $\boldsymbol{p}_{i}$ and its neighbors $ \mathcal{N}(\boldsymbol{p}_{i})$. Thus, Eq.~\eqref{eq:general_conv} becomes
\begin{equation}\label{eq:point_conv}
\boldsymbol{p}'_{i} = \mathcal{G} \Big( \big\{\psi \big( \mathcal{D}(\boldsymbol{p}_{i}, \boldsymbol{p}_{j}) \big) \big\}_{\boldsymbol{p}_{j}\in \mathcal{N}(\boldsymbol{p}_{i})} \Big),
\end{equation}
where $\mathcal{D}(\boldsymbol{p}_{i}, \boldsymbol{p}_{j})$ indicates the encoding function of the geometric association between $\boldsymbol{p}_{i}$ and $\boldsymbol{p}_{j}$. The convolutional output $\boldsymbol{p}'_{i}$ aggregates all the geometric associations between $\boldsymbol{p}_{i}$ and $\mathcal{N}(\boldsymbol{p}_{i})$, thus it could show superior shape awareness.

\vspace{2pt}
\noindent\textbf{Function: Searchable construction.}\,\,~The key problem for the convolution in Eq.~\eqref{eq:point_conv} is how to design the concrete expressions of $\psi(\cdot)$ and $\mathcal{D}(\cdot, \cdot)$. Instead of the common handcrafted manner, we transform this problem into a NAS-like search process. That is, $\mathcal{D}(\cdot, \cdot)$ is devoted to the construction of geometric association encoding while $\psi(\cdot)$ is responsible for all the learnable parameters on this construction.

Technically, we construct $\mathcal{D}(\cdot, \cdot)$ using a directed acyclic graph (DAG). \label{DAG}As the example in Fig.~\ref{Fig3}(b) shows, our DAG can be represented as an ordered sequence of several hidden nodes. Each node is a feature vector and the $i$-th node is denoted as $\boldsymbol{n}_{i}$. To deeply encode the geometric association, a number of searchable essential associations (EA, and EA$_{ij}$ indicates the association between $\boldsymbol{n}_{i}$ and $\boldsymbol{n}_{j}$) are conducted on these hidden nodes. Like deep network with multiple layers, our DAG can also be constructed with multiple levels $\ell$. Furthermore, it is noticeable that there must be a connection between any two nodes (except the final output nodes) in our DAG, and all the connections must go through the essential associations. Therefore, our DAG is capable of sufficiently encoding the geometric association between $\boldsymbol{p}_{i}$ and $\boldsymbol{p}_{j}$. As a result, the whole convolution with our DAG can be searched out in a purely data-driven manner. Note that this search process is different from SGAS \cite{li2020sgas}, which just manually selects seven existing graph convolutions (\textit{e.g.}, EdgeConv~\cite{wang2019dynamic}, GAT~\cite{velivckovic2017graph} and SemiGCN~\cite{kipf2016semi}) as searchable convolution candidates.

\vspace{2pt}
\noindent\textbf{Essential association (EA).}\,\,~The essential association is the core of our DAG. It has the function of transferring the key information from preceding nodes to the output nodes. Instead of elaborate manual design, we propose to search an optimal association for each EA from predefined association candidates. \label{EA}As summarized in Tab.~\ref{tab: EA}, we define five fundamental association candidates after a full investigation in the field of point cloud processing. Due to the capability to learn structural relational features on multiple aspects, the five candidates provide informative enough search space for our DAG in terms of learning geometric associations between $\boldsymbol{p}_{i}$ and its neighbors $\boldsymbol{p}_{j} \in \mathcal{N}(\boldsymbol{p}_{i})$.

\begin{table*}[t]
\setlength{\abovecaptionskip}{0pt}%
\setlength{\belowcaptionskip}{0pt}%
\renewcommand{\arraystretch}{1.0}
\label{sample-table}
\resizebox{1\textwidth}{!}{
\footnotesize
\begin{tabular}{p{3.5cm}llll}
\toprule
Method & $\mathcal{G}{(\cdot)}$ & $\left\{h_{\gamma}\left(\mathcal{D}\left(\boldsymbol{p}_{i}, \boldsymbol{p}_{j}\right)\right)\right\}_{\boldsymbol{p}_{j} \in \mathcal{N}\left(\boldsymbol{p}_{i}\right)}$ & Function expression in PointSeaConv\\
\midrule
PointNet++~\cite{qi2017pointnet++}  & $\operatorname{max}(\cdot)$ & $\operatorname{MLP}(e_2)$ & $\mathop{\operatorname{max}}_{\boldsymbol{p}_{j} \in \mathcal{N}(\boldsymbol{p}_{i})} \{\operatorname{MLP}(e_2)\}$ \\
PointWeb~\cite{zhao2019pointweb}  & $\operatorname{max}(\sum\{\cdot\})$ & $\operatorname{MLP}(e_1, e_3)$ & $\mathop{\operatorname{max}\{\sum}_{\boldsymbol{p}_{j} \in \mathcal{N}(\boldsymbol{p}_{i})} \{\operatorname{MLP}(e_1, e_3)\}\}$\\
DGCNN~\cite{wang2019dynamic} & $\operatorname{max}(\cdot)$  & $\operatorname{MLP}(e_1, e_3)$ & $\mathop{\operatorname{max}}_{\boldsymbol{p}_{j} \in \mathcal{N}(\boldsymbol{p}_{i})} \{\operatorname{MLP}(e_1, e_3)\}$\\
RS-CNN~\cite{liu2019relation}  &  $\operatorname{max}(\cdot)$ & $\operatorname{MLP}(e_1 \oplus e_2 \oplus e_3 \oplus e_4)$ & $\mathop{\operatorname{max}}_{\boldsymbol{p}_{j} \in \mathcal{N}(\boldsymbol{p}_{i})} \{\operatorname{MLP}(e_1 \oplus e_2 \oplus e_3 \oplus e_4)\}$ \\
Pointwise-CNN~\cite{hua2018pointwise}  & $\operatorname{\sum}(\cdot)$  & $\operatorname{MLP}(e_1-e_5)$ & $\mathop{\operatorname{\sum}}_{\boldsymbol{p}_{j} \in \mathcal{N}(\boldsymbol{p}_{i})} \{\operatorname{MLP}(e_1-e_5)\}$\\
\bottomrule
\end{tabular}}
\vspace{0.005cm}
\caption{Several deep learning methods on point clouds can be derived as particular settings of PointSeaConv in Eq.~\eqref{equ3}, by appropriately selecting aggregation function and combinations of essential associations. The definition of $e_1 \sim e_5$ is shown in Tab.~\ref{tab: EA}. $\operatorname{max}$ denotes max pooling and $\sum$ denotes summation.}
\label{tab1}
\vspace{-0.5cm}
\end{table*}

\vspace{2pt}
\noindent\textbf{Parameterization: Searchable convolution.}\,\,~To learn our constructed searchable convolution, we group the learnable parameters in function $\psi(\cdot)$ (Eq.~\eqref{eq:point_conv}) on the DAG into two parts, which are parameterized with $f_{\beta}$ and $h_{\gamma}$. Concretely, $f_{\beta}$ is responsible for the parameters in all the essential associations (EA), which actually determine the construction of DAG. $h_{\gamma}$ is responsible for the parameters in all the multi-layer perceptrons (MLP) on the DAG. Thus the searchable convolution version of Eq. \eqref{eq:point_conv} can be written as
\begin{equation}\label{equ3}
\begin{split}
&\boldsymbol{p}^{\prime}_{i}= \mathcal{G} \Big( \big\{ h_{\gamma} \big( \mathcal{D}(\boldsymbol{p}_{i}, \boldsymbol{p}_{j}) \big) \big\}_{\boldsymbol{p}_{j}\in \mathcal{N}(\boldsymbol{p}_{i})} \Big), \\
&\mathcal{D}(\boldsymbol{p}_{i}, \boldsymbol{p}_{j}) = f_{\beta} \big(\forall \ \text{EA} \in \mathcal{D}(\boldsymbol{p}_{i}, \boldsymbol{p}_{j})\big). \\
\end{split}
\end{equation}
In this way, an optimal combination of essential associations can be searched out to create a suitable convolution, which is capable of sufficiently capturing diverse geometric structures of point clouds.

In implementation, on one hand, both $f_{\beta}$ and $h_{\gamma}$ are shared over each neighboring point $\boldsymbol{p}_{j}\in \mathcal{N}(\boldsymbol{p}_{i})$. Then, with a symmetric aggregation function $\mathcal{G}$, PointSeaConv can be permutation invariant to unordered points while be capable of capturing local structures sufficiently. On the other hand, we adopt k-nearest neighbor approach to acquire $\mathcal{N}(\boldsymbol{p}_{i})$. Hence PointSeaConv can achieve the weight sharing property despite that different local regions are of irregular structures (\textit{i.e.}, variable number of points).

In addition, our searchable convolution shows good generalization in point cloud processing. As summarized in Tab.~\ref{tab1}, most recent convolutions can be seen as special cases of our searchable convolution. For example, DGCNN \cite{wang2019dynamic} can be implemented by configuring MLP with $e_1$ and $e_3$ in Tab.~\ref{tab: EA} to learn geometric associations.

\subsection{Joint Differentiable Optimization Approach}
\label{section3}
\noindent \textbf{Differentiable architecture search.}\,\,~Before introducing our approach, we first briefly review cell-based NAS methods~\cite{liu2018progressive, pham2018efficient, zoph2018learning}. This class of methods represent the architecture as a set of identical cells with different weights, which is represented by directed acyclic graphs (DAG) with an ordered series of nodes. Formally, $x^{(i)}$ denotes the output of the $i$-th node and $(i, j)$ denotes a directed edge from the $i$-th node to $j$-th node. The candidate operations are denoted as $\mathcal{O}$, in which each element $o^{(i, j)}(\cdot)$ propagates the information from $x^{(i)}$ to $x^{(j)}$ across the edge $(i, j)$. In differentiable architecture search methods~\cite{bi2020gold, liu2018darts, xu2019pc}, the continuous relaxation of candidate operations is conducted to obtain the optimal architecture. Consider continuous variables $\alpha=\left\{\alpha^{(i, j)}\right\}$ as architecture parameters for edge $(i, j)$ and the network weights $\omega$, the selection of candidate operations can be relaxed as a softmax mixture over all the possible operations within the operation space $\mathcal{O}$. Then, the output at $j$-th node is the sum of information flows from all its predecessors. Intrinsically, the goal of NAS is to derive the optimal architecture $\alpha^{\star}$ and network weights $\omega^{\star}(\alpha)$ associated with the architecture $\alpha^{\star}$ by solving the following bilevel optimization problem

\begin{equation}
 \begin{split}
\alpha^{\star}=\arg \min _{\alpha} \mathcal{L}_{val}\left(\omega^{\star}(\alpha), \alpha\right),  \\
\text { s.t. } \quad \omega^{\star}(\alpha)=\arg \min _{\omega} \mathcal{L}_{train}\left(\omega, \alpha \right),
  \end{split}
\end{equation}
where $\mathcal{L}_{train}$ and ${\mathcal{L}_{val}}$ indicate the training and validation loss, respectively. After the search process, the final architecture is derived by selecting the path with the highest architecture parameters.

\vspace{2pt}
\noindent \textbf{Joint Optimization.}\,\,~To enable end-to-end training for convolution search, we perform architecture search for the optimal convolution and cell architecture simultaneously under the differentiable architecture search framework as in ~\cite{bi2020gold, liu2018darts, xu2019pc}, denoted as \textit{convolution search} and \textit{network search}, respectively. Intuitively, the overall search framework is shown in Fig.~\ref{Fig3}, which takes a cell structure with 4 nodes and its connection from $x^{(i)}$ to $x^{(j)}$ as an example. Similar to the selection of candidate operations in cell structure search, we define five essential association candidates as search space in convolution search as shown in~Tab.~\ref{tab: EA}, denoted as $\mathcal{E}$. In our framework of joint optimization, in addition to the weights $\omega$ in the network, the whole architectural parameters are denoted as $\rho=\{\theta, \beta\}$, where $\theta$ and $\beta$ indicate parameters of network search and convolution search, respectively. In network search, as the connection from $x^{(i)}$ to $x^{(j)}$, the output of $f^{(i,j)}\left(x^{(i)}\right)$ becomes
\begin{equation}
f^{(i, j)}(x^{(i)})=\sum_{o \in \mathcal{O}} \frac{\exp \left\{\theta_{o}^{(i, j)}\right\}}{\sum_{o^{\prime} \in \mathcal{O}} \exp \left\{\theta_{o^{\prime}}^{(i, j)}\right\}}  o\left(x^{(i)}\right).
\end{equation}
In convolution search, given an input $\boldsymbol{n}_{i}$ and its neighbors $\boldsymbol{n}_{j} \in \mathcal{N}(\boldsymbol{n}_{i})$ in a local neighborhood, the choice of a particular essential association can be relaxed to a softmax mixture in dimension $|\mathcal{E}|$
\begin{equation}
\bar{e}^{(i, j)}(\boldsymbol{n}_{i},\boldsymbol{n}_{j})=\sum_{e \in \mathcal{E}} \frac{\exp \left(\beta_{e}^{(i, j)}\right)}{\sum_{e^{\prime} \in \mathcal{E}} \exp \left(\beta_{e^{\prime}}^{(i, j)}\right)}  e(\boldsymbol{n}_{i},\boldsymbol{n}_{j}).
\end{equation}
Accordingly, the tasks of convolution search and network search can be summarized to learn a set of parameters $\rho=\{\theta, \beta\}$. This bilevel optimization process can be described to updated $\rho$ and $\omega$ alternately
\begin{equation}\label{equ15}
  \begin{split}
\begin{array}{l}
\omega_{t+1} \leftarrow \omega_{t}-\eta_{\omega} \cdot \nabla_{\omega}
\mathcal{L}_{v a l}\left(\omega_t, \rho_t)\right), \\
\rho_{t+1} \leftarrow \rho_{t}-\eta_{\rho} \cdot \nabla_{\rho} \mathcal{L}_{train}\left(\omega_{t+1}, \rho_t)\right),
\end{array}
  \end{split}
\end{equation}
where $\eta_{\omega}$ and $\eta_{{\rho}}$ denote the learning rates for $\omega$ and $\rho$, respectively. For simplicity, we incorporate the parameters of MLP in convolution (denoted by $\gamma$ in Eq.~\eqref{equ3}) into network weights $\omega$.

Notably, the discrete architecture for network search is obtained by retaining each operation with the highest weight, $f^{(i, j)}\left(x^{(i)}\right)=\operatorname{argmax}_{o \in \mathcal{O}} \theta_{o}^{(i, j)}$. With respect to convolution search, we will introduce epsilon-greedy algorithm to reduce the discretization error in the following.

\vspace{2pt}
\noindent \textbf{Epsilon-greedy Algorithm.}\,\,~As pointed in~\cite{bi2020gold, chen2019progressive, tian2020discretization}, the optimization method in DARTS~\cite{liu2018darts} leads to a large discretization error after the search process due to deleting substantial candidate operations with moderate weights. Note that, since convolution search and network search are conducted simultaneously in our method, the risk of discretization error further grows.

To alleviate the discretization error and its accumulation during the search process, we introduce the epsilon-greedy algorithm for efficient optimization of convolution search. First, we make essential association candidates fixed in each step of optimizing the weights $\omega$, where each of them is discretized by selecting the strongest one using greedy algorithm, denoted as $\hat{\beta}_{e}^{(i,j)}=\operatorname{max}_{e^{\prime} \in \mathcal{E}} \beta_{e^{\prime}}^{(i, j)}$. Only in the stage of optimizing ${\beta}_{e}^{(i, j)}$, all the choices of essential association candidates are relaxed. Further, in order to avoid removing all the moderate candidates and reduce the dependence on the parameter initialization, the essential association candidates with the highest weight are selected by a certain probability $\varepsilon$, so that the optimization process of convolution search can be described as
\begin{equation}\label{equ11}
\left\{\begin{array}{l}
P{({\beta}_{e}^{(i, j)}=\hat{\beta}_{e}^{(i, j)}})=1-\varepsilon \\
P({\beta}_{e}^{(i, j)}=\beta_{\text {random }})=\varepsilon
\end{array}\right.,
\end{equation}
where $P(\cdot)$ is a probability distribution of ${\beta}_{e}^{(i, j)}$, $\beta_{\text {random}}$ is a random one-hot vector, and $\varepsilon$ is a hyper-parameter to balance greedy algorithm and random algorithm. Instead of eliminating all weak candidates, epsilon-greedy algorithm retains more candidates that can contribute more or less to training accuracy. In this way, dramatic improvements are achieved by our PointSeaNet on multiple tasks. Detailed settings and analyses are provided in the Sec.~\ref{experiment}.

\section{Experiment}
\label{experiment}
In this section, we conduct comprehensive experiments to demonstrate the capability of PointSeaNet. We first briefly introduce some experimental settings (Sec.~\ref{exp0}). Then, we systematically evaluate PointSeaNet on challenging benchmarks across various point cloud understanding tasks (Sec.~\ref{exp1}). Finally, we provide detailed ablation studies~(Sec.~\ref{exp2}) to validate PointSeaNet thoroughly.

\subsection{Experimental Setting}
\label{exp0}

In our experiment, the cell architecture has $5$ candidate operations: two PointSeaConv, MLP, \textit{skip-connection} and \textit{zero} operation. Each PointSeaConv has $3$ levels with $5$ nodes. Neighboring points are firstly gathered by \textit{k nearest neighbor} in the first operation of each cell. For each cell architecture, there are $6$ nodes with $3$ intermediate nodes, where the output node is defined as the depth-wise concatenation of its four precedents. Each intermediate node must select two input nodes from its precedents. PointSeaNet is obtained through two stages, a search phase and an evaluation phase. During the search phase, $2$ cells with $32$ initial channels and $k=9$ are used to search the optimal cell architecture for $50$ epochs with batch size $16$. SGD is employed with an initial learning rate $0.005$, momentum $0.9$ and weight decay $3 \times 10^{-4}$. A cosine annealing is used to schedule the learning rate with the minimum learning rate $1 \times 10^{-4}$. For the evaluation phase, we stack the searched cell $6$ times and apply $k=20$ to form a larger network, and then the final network is trained from scratch for $400$ epochs with batch size $128$. The Adam optimization algorithm with learning rate $0.001$ and weight decay $1 \times 10^{-4}$ is employed, where a cosine annealing with the minimum learning rate $1 \times 10^{-5}$ is used to schedule the learning rate. We use the same settings for all the NAS methods. Usually, NAS methods use different settings from the handcrafted ones. We report the best results for two class of methods. Regarding the epsilon-greedy algorithm, we set the probability $\varepsilon$ to be $0.5$, which is responsible for balancing the greedy strategy and random strategy when choosing the essential association candidates in convolution search. Furthermore, $\text {PointSeaNet}^{\dagger}$ that omits epsilon-greedy algorithm in PointSeaNet is employed as a baseline of our model. Specifically, our core code is available\footnote{\urllink \texttt{https://github.com/STAR-ALG/PointSeaNet}}.

\begin{table}
\small
\center
\setlength{\tabcolsep}{3.3mm}{
\footnotesize
\begin{tabular}{p{2.7cm}lcc}
\toprule
Method & OA   & \#params & Search Cost \\
\midrule
Pointwise-CNN~\cite{hua2018pointwise}   &86.1  &-    &manual \\
PointNet~\cite{qi2017pointnet}          &89.2  &3.48 &manual \\
PointNet++~\cite{qi2017pointnet++}      &90.7  &1.48 &manual \\
PointCNN~\cite{li2018pointcnn}          &92.2  &0.45 &manual \\
DGCNN~\cite{wang2019dynamic}            &92.2  &1.84 &manual \\
PCNN~\cite{atzmon2018point}             &92.3  &8.10 &manual  \\
PointASNL~\cite{yan2020pointasnl}      &92.9  & - &manual  \\
InterpCNN~\cite{mao2019interpolated}   &93.0  &12.8 &manual  \\
GeoCNN~\cite{lan2019modeling}           &93.4  &- &manual  \\
RS-CNN~\cite{liu2019relation}           &\underline{93.6}  &-  &manual \\
SGAS~\cite{li2020sgas}                  &93.2  &8.49 &0.19\\
\midrule
$\text {PointSeaNet}^{\dagger}$             & 94.0  &6.70 &0.23\\
PointSeaNet              & \textbf{94.2}  &6.75 &0.25 \\
\bottomrule
\end{tabular}}
\vspace{0.1cm}
\caption{Shape classification results (OA: overall accuracy) on ModelNet40.}
\label{tab3}
\vspace{-0.2cm}
\end{table}

\begin{table}[t]
\center
\setlength{\tabcolsep}{3.1mm}{
\footnotesize
\begin{tabular}{p{2.0cm}lcc}
\toprule
Dataset & Method & \#points & mAP(\%)    \\
\midrule
\multirow{6}{*}{ModelNet40} & PointNet~\cite{qi2017pointnet}          & 1k      & 70.5      \\
& PointCNN~\cite{li2018pointcnn}          & 1k          & 83.8   \\
& DGCNN~\cite{wang2019dynamic}            & 1k          & 85.3  \\
& Densepoint~\cite{liu2019densepoint}   &1k  &\underline{88.5}  \\
\cline{2-4}
& $\text {PointSeaNet}^{\dagger}$       &1k  & 89.9    \\
& PointSeaNet            &1k  & $\textbf{90.3}$    \\
\bottomrule
\end{tabular}}
\vspace{0.1cm}
\caption{Shape retrieval results (mAP, \%) on ModelNet40.}
\label{tab6}
\vspace{-0.5cm}
\end{table}

\subsection{PointSeaNet for Point Cloud Processing}
\label{exp1}
\noindent \textbf{Shape classification.}\,\,~We conduct architecture search and evaluation on ModelNet10 and ModelNet40 classification benchmarks~\cite{wu20153d}, respectively. The former contains $3991$ training models and $908$ test models in $10$ classes, and the latter consists of $9843$ training models and $2468$ test models in $40$ classes. 1024 points are uniformly sampled by farthest point sampling. During training, we augment the input data with random anisotropic scaling and translation as in~\cite{klokov2017escape}. During testing, similar to~\cite{qi2017pointnet, qi2017pointnet++}, we conduct ten voting tests with random scaling and average the predictions. Additionally, we do not use normals as additional input.

\begin{figure}[t]
\centering
\subfigure[Cell architecture]{\includegraphics[width=0.85\columnwidth]{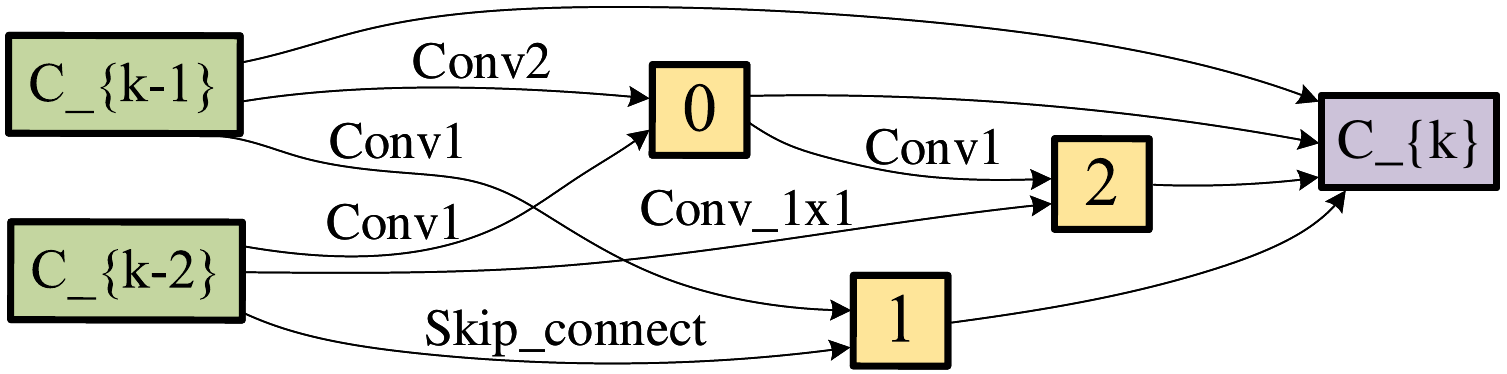}}
\subfigure[Convolution]{\includegraphics[width=0.67\columnwidth]{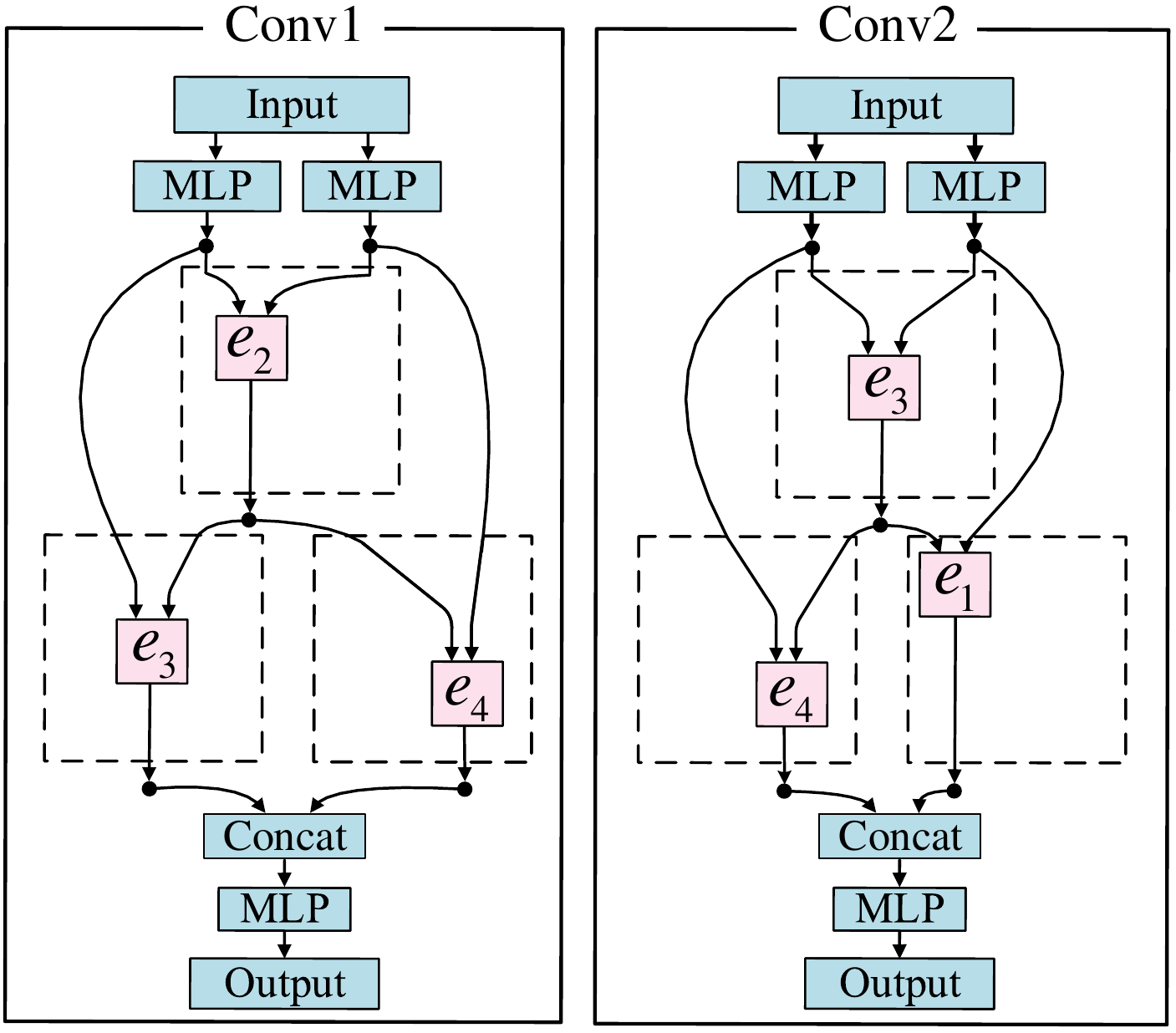}}
\caption{The best architecture and convolution on ModelNet10.}
\label{Fig4}
\vspace{-0.2cm}
\end{figure}

The quantitative comparisons with the other advanced methods are shown in Tab.~\ref{tab3}. Our PointSeaNet outperforms all the other methods, while using only a search cost of $0.23$ GPU day on one NVIDIA TITAN Xp. Compared with SGAS \cite{li2020sgas}, which conducts architecture search with fixed graph convolutions, PointSeaNet reduces the model params by $20.5\%$ and improve the overall accuracy by $1.0\%$. We visualize the searched optimal cell architecture and convolution on ModelNet10 in Fig.~\ref{Fig4}.

\vspace{2pt}
\noindent \textbf{Shape retrieval.}\,\,~To further explore the recognition ability of PointSeaNet, we conduct 3D shape retrieval on ModelNet40. Specifically, we employ the outputs of the penultimate fully-connected layer for shape classification as the global features. We evaluate PointSeaNet on ModelNet40 for this task and uniformly sample $1024$ points as the input. The cosine distance is applied to obtain relative ranking order of each query shapes from the test set. We report \textit{mean Average Precision}~(mAP). Tab.~\ref{tab6} shows the results for the shape retrieval task, where PointSeaNet achieves the best performance with mAP of $90.3\%$ on ModelNet40. Note that, our PointSeaNet surpasses its variant, \textit{i.e.}, $\text {PointSeaNet}^{\dagger}$, with $0.4\%\uparrow$ in mAP.

\begin{table}[t]
\small
\center
\setlength{\tabcolsep}{2.6mm}{
\footnotesize
\begin{tabular}{p{2.3cm}lccc}
\toprule
Method & Input & Class mIou & Instance mIou \\
\midrule
Kd-Net~\cite{klokov2017escape} & 4k & 77.4 & 82.3 \\
PointNet~\cite{qi2017pointnet} & 2k & 80.4 & 83.7 \\
PCNN~\cite{atzmon2018point} & 2k & 81.8 & 85.1  \\
PointNet++~\cite{qi2017pointnet++} & 2k, nor & 81.9 & 85.1  \\
SyncCNN~\cite{yi2017syncspeccnn} & mesh & 82.0 & 84.7  \\
SPLATNet~\cite{su2018splatnet} & - & 82.0 & 84.6 \\
DGCNN~\cite{wang2019dynamic} & 2k & 82.3 & 85.1  \\
RS-CNN~\cite{liu2019relation} & 2k & 84.0 & 86.2  \\
Densepoint~\cite{liu2019densepoint} & 2k & \underline{84.2} & \underline{86.4}  \\
\midrule
$\text {PointSeaNet}^{\dagger}$  & 2k  & 85.2 & 87.3 \\
PointSeaNet          & 2k  & \textbf{85.7} & \textbf{87.8} \\
\bottomrule
\end{tabular}}
\vspace{0.1cm}
\caption{Shape part segmentation results (\%) on ShapeNetPart (nor: normal, `-': unknown).}
\label{tab4}
\vspace{-0.2cm}
\end{table}

\begin{table}
\center
\setlength{\tabcolsep}{3.7mm}{
\footnotesize
\begin{tabular}{lcccc}
\toprule
\multirow{2}*{Method} & \multicolumn{2}{c}{Area-5} & \multicolumn{2}{c}{6-fold}\\
\cmidrule(r){2-3}  \cmidrule(r){4-5}
~ & OA & mIoU & OA & mIoU\\
\midrule
PointNet~\cite{qi2017pointnet} & - & 41.1 & 78.6 & 47.6 \\
PointNet++~\cite{qi2017pointnet++} & - & - & 81.0  & 54.5  \\
DGCNN~\cite{wang2019dynamic} & - & - & 84.1 & 56.1  \\
PointCNN~\cite{li2018pointcnn} & 85.9 & 57.3 & 88.1 & 65.4  \\
LSANet~\cite{chen2019lsanet} & - & - & 86.8 & 62.2 \\
SPG~\cite{landrieu2018large} & 86.4 & 58.0 & 85.5 & 62.1  \\
RandLA-Net~\cite{hu2020randla} & - & - & 87.2 & 68.5 \\
PAG~\cite{pan2020pointatrousgraph} & 86.8 & 59.3 & 88.1 & 65.9 \\
PointSIFT~\cite{jiang2018pointsift} & - & - & \underline{88.7}  & 70.2 \\
Pointweb~\cite{zhao2019pointweb} & 87.0 & 60.3 & 87.3 & 66.7  \\
HPEIN~\cite{jiang2019hierarchical} & \underline{87.2} & 61.9 & 88.2 & 67.8  \\
KPConv~\cite{thomas2019kpconv} & - & \underline{67.1} & - & \underline{70.6}  \\
\midrule
$\text {PointSeaNet}^{\dagger}$  & 88.1 & 68.2  & 89.6 & 71.2 \\
PointSeaNet   & \textbf{89.2} & \textbf{69.0}  & \textbf{90.3} & \textbf{71.9} \\
\bottomrule
\end{tabular}}
\vspace{0.1cm}
\caption{Scene segmentation results (\%) on S3DIS (`-': unknown).}
\label{tab5}
\vspace{-0.5cm}
\end{table}

\vspace{2pt}
\noindent \textbf{Shape part segmentation.}\,\,~For this task, we search the optimal convolution and cell architecture using stacked identical cells on the ShapeNetPart benchmark~\cite{yi2016scalable}, and then the searched cell is stacked to form a larger network, which is retrained on ShapeNetPart. ShapeNetPart consists of $16881$ shapes with $16$ categories, which is labeled in $50$ parts in total. Following~\cite{qi2017pointnet}, we randomly sample $2048$ points as the input and concatenate the one-hot encoding of the object label into the last feature layer. During testing, we also perform ten voting tests using random scaling. Evaluation metrics contain two types of mIoU that are averaged across all classes and all instances respectively. Tab.~\ref{tab4} gives the results in this experiment, where PointSeaNet outperforms the best handcrafted method, Densepoint~\cite{liu2019densepoint}, with $1.5\uparrow$ in class mIoU and $1.4\uparrow$ in instance mIoU respectively. The dramatic improvements validate the capability of our method to learn fine-grained features.

\vspace{2pt}
\noindent \textbf{Large-scale scene segmentation.}\,\,~In this experiment, we perform 3D scene segmentation to evaluate our PointSeaNet on the S3DIS~\cite{armeni20163d} benchmark. As a large-scale public dataset, the S3DIS dataset contains $271$ million points belonging to 6 large-scale indoor areas with $13$ classes. We conduct the optimal convolution and cell architecture search on the S3DIS~\cite{armeni20163d} benchmark. To adequately measure the generalization ability of our PointSeaNet, we adopt both Area-$5$ and standard $6$-fold cross validation as test setting. As listed in Tab.~\ref{tab5}, our PointSeaNet outperforms other advanced methods. Specifically, PointSeaNet achieves a superior results compared with its variant $\text {PointSeaNet}^{\dagger}$, with $0.7\uparrow$ in overall accuracy and $0.7\uparrow$ in mIoU with the standard 6-fold cross validation as test setting.

\noindent \textbf{Normal estimation.}\,\,~We evaluate PointSeaNet with the same parameters as in the shape part segmentation. The optimal architecture are searched on ModelNet40, and then a larger network composed of searched cells is evaluated with normal estimation as a supervised regression task on ModelNet40. $1024$ points are uniformly sampled as the input. The cosine-loss between the normalized output and the normal ground truth is used to train PointSeaNet. The results in Tab.~\ref{tab7} show that PointSeaNet outperforms all the compared methods with a lower error of $0.10$, which significantly reduces the error of RS-CNN~(0.15) by $33.3\%$.

\begin{table}[t]
\label{sample-table}
\center
\setlength{\tabcolsep}{3.6mm}{
\footnotesize
\begin{tabular}{p{2cm}lccc}
\toprule
Dataset & Method  & \#points & error  \\
\midrule
\multirow{6}{*}{ModelNet40} & PointNet ~\cite{qi2017pointnet}   &1k     & 0.47            \\
                        & PointNet++~\cite{qi2017pointnet++}    &1k     & 0.29   \\
                        & PCNN~\cite{atzmon2018point}    &1k     & 0.19   \\
                        & MC-Conv ~\cite{hermosilla2018monte}   &1k     & 0.16   \\
                        & RS-CNN~\cite{liu2019relation}    &1k     & \underline{0.15}   \\
                        & Densepoint~\cite{liu2019densepoint}  &1k   & \underline{0.15}  \\
                        \cline{2-4}
                        & $\text {PointSeaNet}^{\dagger}$   & 1k & 0.12  \\
                        & PointSeaNet           & 1k & \textbf{0.10}  \\
\bottomrule
\end{tabular}}
\vspace{0.1cm}
\caption{Normal estimation error on ModelNet40.}
\label{tab7}
\vspace{-0.5cm}
\end{table}

\subsection{Ablation study}
\label{exp2}

\noindent \textbf{Sensitivity to hyperparameters.}\,\,~We conduct experiments to evaluate the sensitivity of our method to hyperparameters, \textit{i.e.}, upon different settings of the number of cells, PointSeaConv and DAG levels. As shown in Tab.~\ref{tab8}, PointSeaNet can get a decent accuracy of $93.7\%$ with only $3$ cells, $2$ PointSeaConv and $2$ DAG levels. Note that, our PointSeaNet with $2$ DAG levels, $2$ PointseaConv and $6$ cells achieves the best performance, instead of the version with the largest amount of parameters. This clearly indicates that deeper level can improve performance to some extent, yet the success of PointSeaNet does not entirely come from introducing more parameters.

\vspace{2pt}
\noindent \textbf{Analysis of essential associations.}\,\,~We experiment on ModelNet40 for shape classification to evaluate the searched architecture, to analyse the five essential associations~(Tab.\ref{tab: EA}). The results in Tab.~\ref{tab9} show that the baseline~(model A) gets a low accuracy of $81.3\%$, which is set to architecture search with only $e_{1}$. Yet with local features denoted by $e_{2}$, it is significantly improved to $87.1\%$~(model B), which shows that local features are crucial for improve performance. Then, when using geometric relations $e_{3}$ to enhance the representation ability of PointSeaNet, the accuracy can be further improved to $90.2\%$~(model D). Noticeably, Euclidean distance $e_{4}$ can bring a boost of $2.7\%$~(model E). Finally, the salient information $e_{5}$ can result in an accuracy variation of $1.1\%$~(model F).

\vspace{2pt}
\noindent \textbf{Effectiveness of convolution search and epsilon-greedy algorithm.}\,\,~We provide a detailed analysis to better understand the contributions of PointSeaNet. As can be seen in Tab.~\ref{tab3}, even without the epsilon-greedy algorithm, $\text {PointSeaNet}^{\dagger}$ can also achieve a superior result~($94.0\%$) compared with the state-of-the-art handcrafted method RS-CNN~\cite{liu2019relation}~($93.6\%$) and the best point-cloud-NAS method SGAS~\cite{li2020sgas}~($93.2\%$). Though equipped with differentiable architecture search framework, SGAS adpots the existing graph convolutions, leading to restrict its capability to sufficiently capture geometric structures on point clouds. This adequately validate the effectiveness of our searchable convolution. Furthermore, the epsilon-greedy algorithm we introduce into NAS can significantly boost performance for various point cloud analysis tasks. As shown in Tab.~\ref{tab6}, PointSeaNet surpasses its variant that omits the epsilon-greedy algorithm, \textit{i.e.}, $\text {PointSeaNet}^{\dagger}$, with $0.4\uparrow$ in mAP on ModelNet40 shape retrieval. Regarding large-scale scene segmentation on S3DIS, the results in Tab.~\ref{tab5} show that PointSeaNet brings $0.9\%$ overall accuracy gains and $0.8\%$ mIoU gains over $\text {PointSeaNet}^{\dagger}$ with Area-5 as test scene.

\begin{table}[t]
\label{sample-table}
\center
\setlength{\tabcolsep}{1.8mm}{
\footnotesize
\begin{tabular}{ccccc}
\toprule
\# Cells & \# PointSeaConv & \# DAG levels & \# params(M) & OA(\%) \\
\midrule
3 &2  &2  &4.15 &93.7 \\
6 &2  &2  &6.75 &\textbf{94.2} \\
9 &2  &2  &9.36 & 94.1 \\
\midrule
6 &1  &2 &6.26 &93.8\\
6 &2  &2 &6.75 & \textbf{94.2}\\
6 &3  &2 &6.95 & 94.0\\
\midrule
6 &2  &1  &6.53 &93.5\\
6 &2  &2  &6.75 & \textbf{94.2}\\
6 &2  &3  &8.72 & 93.9\\
\bottomrule
\end{tabular}}
\vspace{0.1cm}
\caption{The comparisons of different number of cells, PointSeaConv and DAG levels during the evaluation phase.}
\vspace{-0.5cm}
\label{tab8}
\end{table}

\begin{table}[t]
\center
\setlength{\tabcolsep}{2.65mm}{
\footnotesize
\begin{tabular}{cccccccc}
\toprule
Model  & $e_1$ & $e_2$ & $e_3$ & $e_4$ & $e_5$ & $\epsilon$-greedy & OA(\%) \\
\midrule
A     &\checkmark  &    &           &           &                    &            &81.3 \\
B     &\checkmark  &\checkmark  &  &           &                     &           &87.1 \\
C    &\checkmark  &  &\checkmark  &  &                     &           &88.9\\
D    &\checkmark  &\checkmark  &\checkmark  &  &                     &           &90.2\\
E    &\checkmark  &\checkmark  &\checkmark  &\checkmark &             &           &92.9 \\
F    &\checkmark  &\checkmark  &\checkmark  &\checkmark &\checkmark                         &           & 94.0 \\
G    &\checkmark  &\checkmark  &\checkmark  &\checkmark &\checkmark              &\checkmark                     & \textbf{94.2} \\
\bottomrule
\end{tabular}}
\vspace{0.1cm}
\caption{The comparisons of choices of epsilon-greedy algorithm and several essential associations during the evaluation phase. The definition of $e_{1} \sim e_{5}$ is shown in Tab.~\ref{tab: EA}.}
\label{tab9}
\vspace{-0.57cm}
\end{table}

\section{Conclusion}
In this work, we present PointSeaConv, a differentiable convolution search paradigm that operates on point clouds. PointSeaConv is capable of creating an optimal convolution to sufficiently learn local structural features in a purely data-driven manner. For this purpose, a dynamic and learnable directed acyclic graph (DAG) is constructed to represent the whole convolution. Then a joint and differentiable optimization framework is developed to search for core convolution and external architecture. Meanwhile, by incorporating epsilon-greedy algorithm into convolution search, the discretization error is sharply alleviated during the search process, resulting in remarkably better performance.

{\small
\bibliographystyle{ieee_fullname}
\bibliography{egbib}
}

\end{document}